\documentclass[environsciproc,article,submit,pdftex,moreauthors]{Definitions/mdpi} 

\preto{\abstractkeywords}{\nolinenumbers}

\firstpage{1} 
\pubvolume{1}
\issuenum{1}
\articlenumber{0}
\pubyear{2023}
\copyrightyear{2023}
\datereceived{ } 
\daterevised{ } 
\dateaccepted{ } 
\datepublished{ } 
\hreflink{https://doi.org/} 

\usepackage{afterpage}
\usepackage{graphicx}
\usepackage{subcaption}
\usepackage{comment}

\Title{Trainable Noise Model as an XAI evaluation method: application on Sobol for remote sensing image segmentation}

\TitleCitation{Trainable Noise Model as an XAI evaluation method: application on Sobol for remote sensing image segmentation}

\Author{Hossein Shreim$^{1,2}$, Abdul Karim Gizzini$^{3}$ and Ali J. Ghandour\orcidA{} $^{2}$*}

\AuthorNames{Hossein Shreim, Abdul Karim Gizzini and Ali J. Ghandour}

\AuthorCitation{Shreim, H.; Gizzini, A.; Ghandour, A.}

\address{%
$^{1}$ \quad Lebanese University;\\
$^{2}$ \quad National Center for Remote Sensing - CNRS, Lebanon;\\
$^{3}$ \quad Center for Digital Systems, IMT Nord Europe, Institut Mines-Télécom, University of Lille;}

\corres{Corresponding author: Ali J. Ghandour, aghandour@cnrs.edu.lb;}

\abstract{eXplainable Artificial Intelligence (XAI) has emerged as an essential requirement when dealing with mission-critical applications, ensuring transparency and interpretability of the employed black box AI models. The significance of XAI spans various domains, from healthcare to finance, where understanding the decision-making process of deep learning algorithms is essential. Most AI-based computer vision models are often black boxes; hence, providing the explainability of deep neural networks in image processing is crucial for their wide adoption and deployment in medical image analysis, autonomous driving, and remote sensing applications.
Existing XAI methods aim to provide insights about the methodology used by the black-box model in making decisions by highlighting the most relevant regions within the input image that contribute to the model's prediction. Recently, several XAI methods for image classification tasks have been introduced. In contrast, image segmentation has received comparatively less attention in the context of explainability, although it is a fundamental task in computer vision applications, especially in remote sensing. Only some research proposes gradient-based XAI algorithms for image segmentation.
This paper adapts the recent gradient-free  Sobol XAI method for semantic segmentation. To measure the performance of the Sobol method for segmentation, we propose a quantitative XAI evaluation method based on a learnable noise model. The main objective of this model is to induce noise on the explanation maps, where a higher induced noise signifies low accuracy and vice versa. A benchmark analysis is conducted to evaluate and compare performance of three XAI methods, including Seg-Grad-CAM, Seg-Grad-CAM++ and Seg-Sobol using the proposed noise-based evaluation technique. This constitutes the first attempt to run and evaluate XAI methods using high-resolution satellite images. Our code is publicly available at this \href{https://github.com/geoaigroup/GEOAI-ECRS2023}{Repo}.}

\keyword{explainable artificial intelligence (XAI); remote sensing; XAI Evaluation; Semantic segmentation}

\begin{document}

\section{Introduction}
Deep neural networks have achieved remarkable success in various computer vision tasks such as classification, detection, and semantic segmentation. However, they lack interpretability because of their black-box-based processing. Consequently, explainable artificial intelligence (XAI) is a crucial need to understand and interpret the decisions made by any deep learning black box model. Numerous XAI methods have been proposed~\cite{jung2021towards, fel2021look, koker2021u} to provide valuable insights into the inner workings of the model and help build trust and confidence in its decision-making process. Generally speaking, XAI methods for image processing tasks provide explanations as saliency maps that highlight the most influential regions of the input that contribute significantly to the model's prediction. The most recent XAI methods are dedicated to classification tasks, where XAI for segmentation is still largely unexplored. There are two main categories of XAI methods~\cite{nielsen2022robust}: (\textit{i}) perturbation-based, where the concept is to perturb input features and record the effect of these changes on model performance without diving into the internal architecture of the considered model, and (\textit{ii}) gradient-based methods where the gradients of the output are calculated with respect to the extracted features or the input via backpropagation and used to estimate attribution scores. We note that internal access to the model architecture is essential in these methods.

Motivated by the fact that evaluating the performance and reliability of XAI methods is crucial to determine their efficiency and reliability for real-world applications. In this work, we propose a quantitative XAI evaluation approach that facilitates a deeper understanding of the performance of any XAI method. The proposed XAI evaluation approach is based on the methodology of the U-Noise model~\cite{koker2021u} that was initially used as an XAI method. The original U-Noise aims to interpret a pre-trained segmentation model by employing an external model that is responsible for adding noise to the input image without harming the accuracy of the pre-trained model. By doing this, the U-Noise model defines the most important pixels contributing towards the target class segmentation as those assigned low noise weights.

In this context, our proposed evaluation methodology is to feed the XAI saliency map multiplied by the input image to the U-Noise model. Therefore, the U-Noise model serves as a tool for assessing and quantifying the fidelity of XAI methods by adding noise to the important highlighted pixels. Inspired by the recent work proposed in~\cite{vinogradova2020towards}, where the gradient-weighted class activation mapping (Grad-CAM) XAI method has been adapted from the classification task to the segmentation task. In this work, we adapt the recently proposed perturbation-based Sobol method~\cite{fel2021look} to segmentation. Rather than calculating the Sobol indices for a single classification output as performed in the original work~\cite{fel2021look}, we calculated the Seg-Sobol indices with respect to multiple values of the segmentation output mask considering a specific target class.

To demonstrate the effectiveness of our proposed evaluation technique, we performed experiments on \hl{two datasets, Cityscapes dataset}~\cite{Cordts2016Cityscapes} \hl{which contains a diverse set of semantic urban scene labels, and WHU dataset which contains satellite images focusing on roof buildings segmentation}~\cite{nasrallah2022lebanon}. Our experimental results demonstrate the ability of the proposed evaluation technique to compare the fidelity of different XAI methods, enabling a more comprehensive and objective assessment of any XAI method. To sum up, the contribution of this paper is threefold:

\begin{itemize}

    \item Propose a quantitative XAI evaluation approach using a learnable noise model. Our evaluation methodology is based on feeding the saliency map combined with the input image to the noise model. Then, on the basis of the generated noise mask, statistical metrics are computed to quantitatively evaluate the performance of any XAI method.

    \item Adapt the recently proposed perturbation-based Sobol XAI method from classification to semantic segmentation.
        
    \item Benchmark the performance of the adapted Sobol with the gradient-based XAI methods Seg-Grad-CAM and Seg-Grad-CAM++ using the WHU dataset for building footprint segmentation.
    
\end{itemize}

\begin{figure}[t]
\centering
\includegraphics[width=1\textwidth]{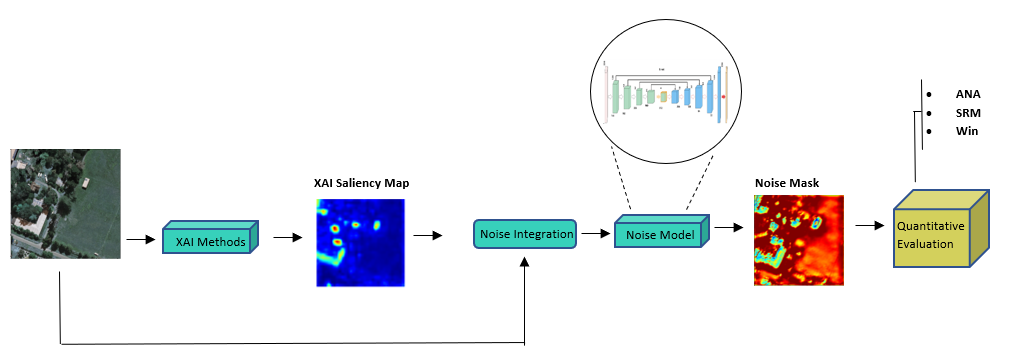}
\caption{Propoosed quantatitave evaluation of XAI methods using U-Noise model.}
\label{fig:noise_model}
\end{figure}

\section{Proposed Trainable Noise Model XAI Evaluation}

\subsection{Methodology}
The saliency map of the XAI method assumes that the highlighted pixels contribute more to the model decision. To validate whether the highlighted pixels are really relevant to the model decision, XAI evaluation is a must. In this context, our proposed XAI evaluation approach is based on combining the saliency map generated by a specific XAI method with the original image and then feeding the resultant mask, denoted as the explanation map, to a trained U-Noise model. The U-Noise model is responsible for adding noise to the explanation map. A better XAI method would receive less added noise, as it retains the correct important pixels that contribute to the model decision. Figure~\ref{fig:noise_model} illustrates the block diagram of the proposed U-Noise XAI evaluation approach. 

In order to achieve a comprehensive evaluation analysis of XAI, the explanation maps are generated according to the following methodology.

Given an original image $\boldsymbol{I}$ and its corresponding saliency map $\boldsymbol{L}_{c}$ generated by an XAI method where $c$ denotes the target class, the explanation map $\boldsymbol{I}^{\prime}$ can be manipulated as follows:

\begin{enumerate}
    \item \textbf{Multiplication}:  The original input image is directly multiplied by the saliency map, highlighting regions of the image assumed important by the XAI method as shown in Equation~\ref{eq:mul}:
    
    \begin{equation}
    \boldsymbol{I}^{\prime}_{\text{mul}} = \boldsymbol{I} \times \boldsymbol{L}_{c}.
    \label{eq:mul}
    \end{equation}
   
    \item \textbf{Addition}: By adding the saliency map to the original image, we augment the image with importance scores, potentially highlighting regions of interest, as shown in Equation~\ref{eq:add}:
    
    \begin{equation}
    \boldsymbol{I}^{\prime}_{\text{add}} = \boldsymbol{I} + \boldsymbol{L}_{c}.
    \label{eq:add}
    \end{equation}
    
    \item \textbf{Normal Sampling with Multiplication}: Similar to the "Normal Sampling with Addition" method, but with multiplication instead of addition. This emphasizes or de-emphasizes regions based on the importance scores and the sampled noise as shown in Equation~\ref{eq:nsamplingmultiplication}:
    \begin{equation}
   \boldsymbol{I}^{\prime}_{\text{nsm}} = \boldsymbol{I} \times N(\boldsymbol{L}_{c}).
   \label{eq:nsamplingmultiplication}
    \end{equation}

    \item \textbf{Normal Sampling with Addition}: To introduce variability in the pixels of the explanation map, $\boldsymbol{L}_{c}$ is sampled from a normal distribution. The resulting sampled values are then added to the original image, as shown in Equation~\ref{eq:nsamplingaddition}:
 
    \begin{equation}
    \boldsymbol{I}^{\prime}_{\text{nsa}} = \boldsymbol{I} + N(\boldsymbol{L}_{c}).
    \label{eq:nsamplingaddition}
    \end{equation}
    
\end{enumerate}

\begin{figure}[t]
\centering
\includegraphics[width=\textwidth]{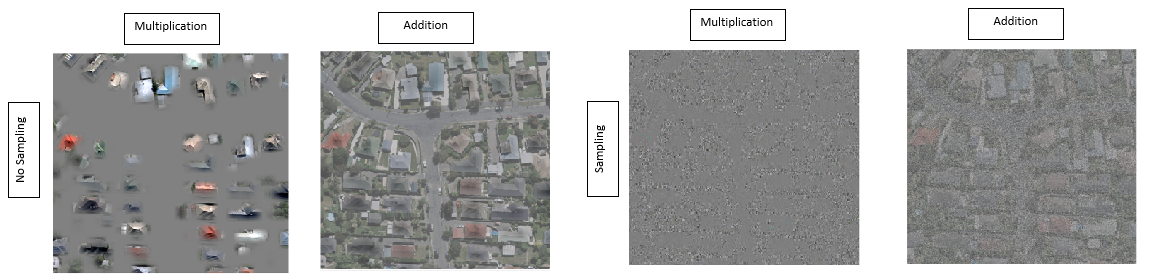}
\caption{Different Integration Techniques.}
\label{fig:I_generation}
\end{figure}

Figure~\ref{fig:I_generation} illustrates the proposed explanation map generation methods. We can clearly notice the impact of each method on generating the explanation map. The use of normal sampling with multiplication (Equation~\ref{eq:nsamplingmultiplication}) is expected to not give a reasonable evaluation, as the U-Noise model was not trained on images with such a distribution. For the scope of this work, we will mainly rely on the Multiplication method with no sampling introduced in Equation~\ref{eq:mul}.

\subsection{Metrics}

In this work, we propose the following two metrics in order to quantitatively report the results of the U-Noise model:

\begin{enumerate}
    \item \textbf{Average Noise Added (ANA)}: This metric computes the mean value of the output of the U-noise model denoted by $\boldsymbol{O} \in \mathbb{R}^{u \times v} $. A higher $ANA$ indicates that the XAI method introduced more noise to the input image, \hl{which means the lower this metric is, the better}.
    \begin{equation}
    ANA = \frac{1}{N} \sum_{(u,v)} O_{i,j},~~~ N = uv.
    \label{eq:ana}
    \end{equation}
    
    \item \textbf{Second Raw Moment (SRM)}: This metric represents the variance of the noise distribution. A higher $SRM$ suggests that the noise introduced by the trained noise model is spread more away from zero \hl{which also means that the lower this metric is, the better}.
    \begin{equation}
    SRM = \frac{1}{N} \sum_{(u,v)}  (N_{i,j})^2
    \label{eq:srm}
    \end{equation}
\end{enumerate}

\begin{figure}[t]
\centering
\includegraphics[width=1\textwidth]{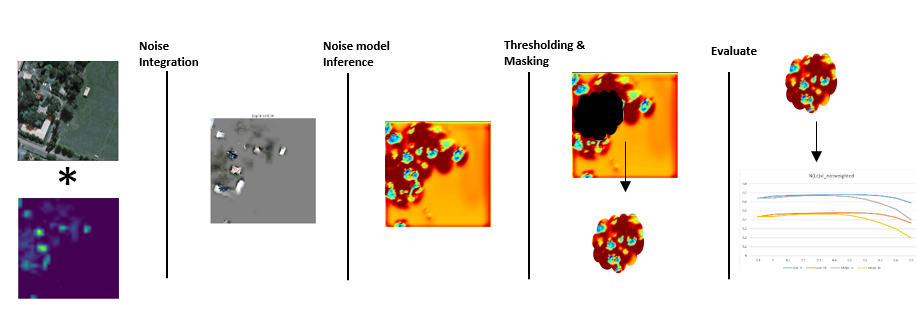}
\caption{Thresholding operation as an additional step to overcome gray areas effect; We first  integrate saliency map of XAI method with original image. Then, we run inference through the Noise model, and apply thresholding before we calculate the evaluation metrics.}
\label{fig:thre}
\end{figure}

\section{Results}
This section presents a quantitative evaluation of the U-Noise-based XAI evaluation method using the Cityscapes and WHU datasets.

\subsection{Cityscapes}
The utility model used was trained to segment the $Road$ class of the Cityscapes dataset. It is worth mentioning that to efficiently evaluate the benchmarked XAI methods, a thresholding operation should be applied to the generated noise mask. This is due to the presence of gray regions within the explanation map, as illustrated in Figure~\ref{fig:thre}.

\begin{figure}[t]
\centering
\includegraphics[width=\textwidth]{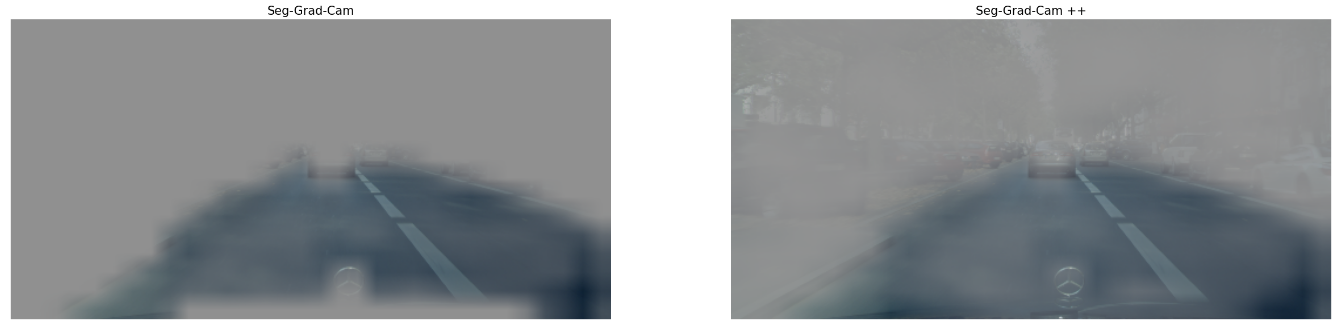}
\caption{(a) Saliency Maps for Seg-Grad-CAM and (b) Saliency Maps for Seg-Grad-CAM++, using Equation~\ref{eq:mul} (multipliclation with no sampling integration technique) over a sample image of Cityscapes dataset.}
\label{fig:I_generation_cityscapes}
\end{figure}

\begin{figure}[t]
\centering
\includegraphics[width=\textwidth]{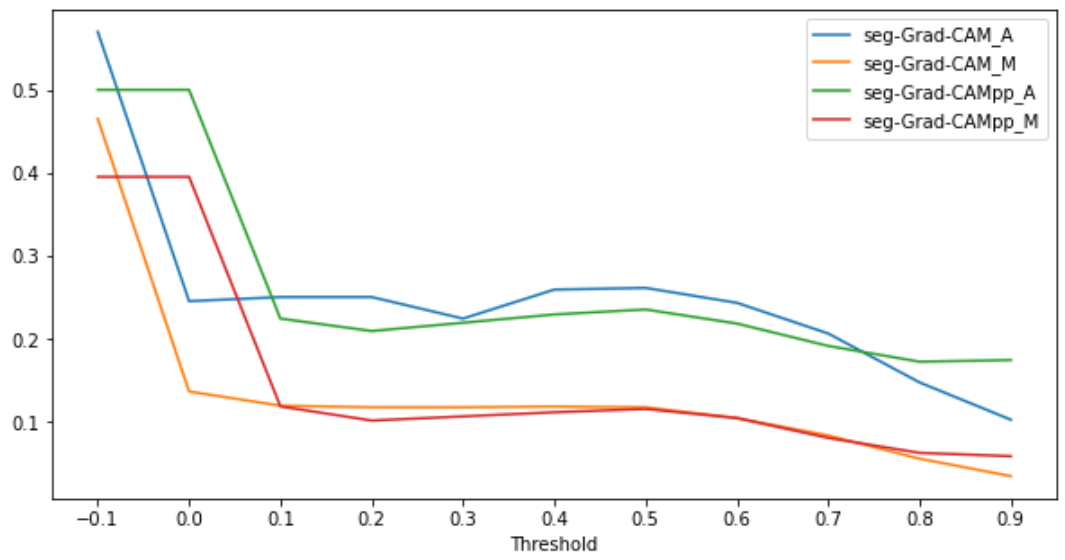}
\caption{Results for the two benchmarked XAI methods over different threshold values: Seg-Grad-CAM\_A and Seg-Grad-CAM++\_A are the average noise added on Seg-Grad-CAM and Seg-Grad-CAM++, respectively. Seg-Grad-CAM\_M and Seg-Grad-CAM++\_M are the second raw moment for noise added on Seg-Grad-CAM and Seg-Grad-CAM++, respectively.}
\label{fig:graph_cityscapes}
\end{figure}

Figure~\ref{fig:I_generation_cityscapes} shows the saliency maps of Seg-Grad-CAM~\cite{vinogradova2020towards} and Seg-Grad-CAM++~\cite{gradcampp}, multiplied by the original image. Figure~\ref{fig:graph_cityscapes} shows the average and second raw moment of the added noise mask for the two XAI methods compared, where the x-axis corresponds to the masking threshold and the y-axis represents the metrics $ANA$ and $SRM$, introduced in Equations {\ref{eq:ana}} and {\ref{eq:srm}}. Starting with a threshold of -0.1, which dictates that no thresholding was performed, the evaluation metrics were calculated on the entire noise mask. Seg-Grad-CAM++ shows lower $ANA$ and $SRM$ than Seg-Grad-CAM, indicating that Seg-Grad-CAM++ provides a better explanation of the utility model, which is consistent with the literature. 

\subsection{WHU}
using the WHU dataset, we benchmark two recent gradient-based XAI methods, Seg-Grad-CAM and Seg-Grad-CAM++, in addition to our adapted Seg-Sobol method.

\begin{figure}[t]
  \centering
  \includegraphics[width=\textwidth]{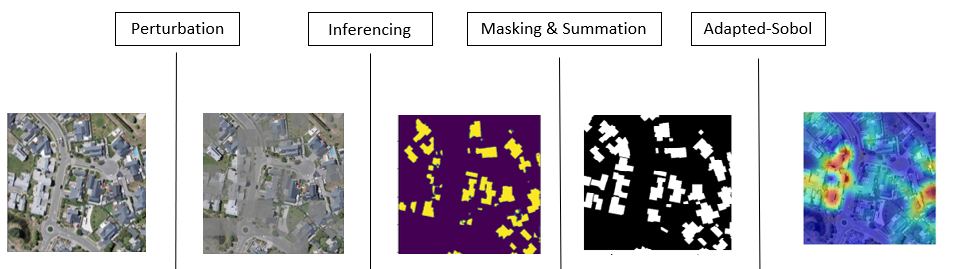}
  \caption{Seg-Sobol: Adaptation of Sobol method from classification to segmentation.}
  \label{fig:sobol_adapt}
\end{figure}

Sobol XAI method~\cite{fel2021look} was initially developed for classification models, where the idea is to perturb the image with several noisy masks and calculate the Sobol indices for each input feature with respect to the output of the classification model taking into account the applied perturbation. The calculated Sobol indices reflect the impact of the applied perturbations on the prediction of the black-box model. For semantic segmentation, the Sobol indices should be calculated with respect to the summation of target-class pixels within the output probability mask. Sobol has the advantage of not needing to have access to the model's internal architecture. Figure~\ref{fig:sobol_adapt} shows the steps to adapt the Sobol method to semantic segmentation, which we refer to as Seg-Sobol.

\begin{figure}[h]
  \centering
  \includegraphics[scale=0.4]{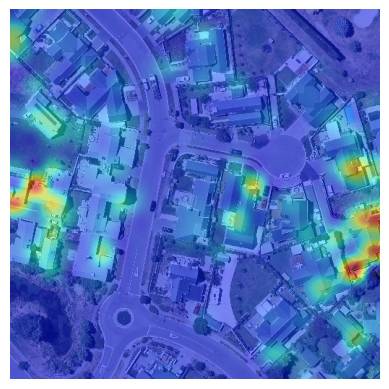}
  \caption{Seg-Sobol results with grid size = 11 using sample from the WHU dataset.}
  \label{fig:seg-sobol result}
\end{figure}

Seg-Sobol saliency map highlights the building surroundings with different intensities as important regions in segmenting building pixels. The results in Figure~\ref{fig:seg-sobol result} are qualitatively plausible, where the highlighted buildings and regions are thought to be important for the segmentation process.

\begin{figure}[t]
  \centering
  \includegraphics[width=\textwidth]{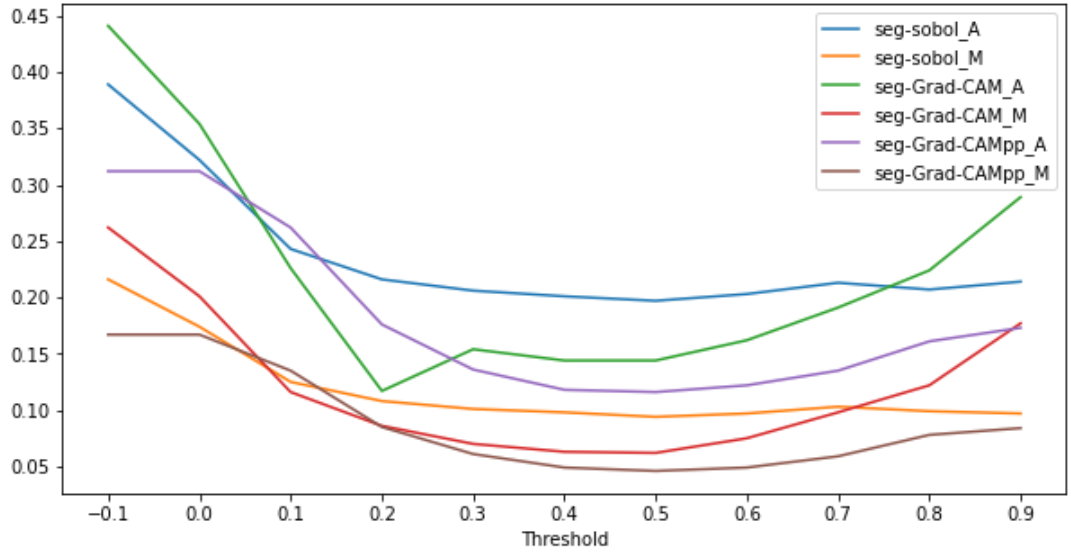}
  \caption{Quantitative metrics results for the benchmarked XAI methods using Equation~\ref{eq:mul} (multiplcation with no sampling) over different threshold values. Seg-Sobol\_A, Seg-Grad-CAM\_A and Seg-Grad-CAM++\_A are the average noise added on Seg-Sobol, Seg-Grad-CAM, and Seg-Grad-CAM++, respectively. Seg-Sobol\_M, Seg-Grad-CAM\_M and Seg-Grad-CAM++\_M are the second raw moment for noise added on Seg-Sobol, Seg-Grad-CAM, and Seg-Grad-CAM++, respectively.}
  \label{fig:graph}
\end{figure}

Figure~\ref{fig:graph} shows the average and the second raw moment of the added noise mask for the three benchmarked XAI methods, \hl{where the x-axis corresponds to the masking threshold and the y-axis represents $ANA$ and $SRM$ metrics, introduced in Equations {\ref{eq:ana}} and {\ref{eq:srm}}}. Seg-Grad-CAM++ shows the lowest noise average, followed by Seg-Sobol and Seg-Grad-CAM. This is also the case with the second raw moment metric. The same results are also observed for the threshold value of 0. For threshold = 0.1, Seg-Grad-CAM receives the lowest noise average and thus outperforms the other two methods. Future work will investigate means to improve the Seg-Sobol explanation outcome for earth observation segmentation use cases.

\section{Conclusions}
In our research, we successfully adapted the Sobol XAI method to better understand image segmentation tasks. To evaluate its effectiveness, we introduced a unique noise model technique. When we compare Seg-Sobol with other methods such as Seg-Grad-CAM and Seg-Grad-CAM++, it showed promising results. Furthermore, using high-resolution satellite images for our tests was a new and important step. These findings are crucial because they make AI-driven earth observation applications more transparent and easier to understand, paving the way for safer and more reliable real-world applications.

\vspace{6pt} 



\authorcontributions{
Conceptualization, Hossein Shreimm Abdul Karim Gizzini, and Ali J. Ghandour; Data curation, Hossein Shreim; Formal analysis, Hossein Shreim, Abdul Karim Gizzini, and Ali J. Ghandour; Investigation, Hossein Shreim; Methodology, Abdul Karim Gizzini, Hossein Shreim; Project administration, Ali J. Ghandour; Resources, Ali J. Ghandour; Software, Hossein Shreim; Supervision, Abdul Karim Gizzini and Ali J. Ghandour; Validation Hossein Shreim, Abdul Karim Gizzini, and Ali J. Ghandour; Visualization,Hossein Shreim; Writing - original draft, Hossein Shreim; Writing - review \& editing,  Abdul Karim Gizzini, and Ali J. Ghandour;
}

\funding{``This research received no external funding''.}

\conflictsofinterest{``The authors declare no conflict of interest.''} 


\begin{adjustwidth}{-\extralength}{0cm}

\reftitle{References}




\bibliography{reference}

\begin{thebibliography}{999}

\bibitem[Jung and Oh(2021)]{jung2021towards}
Jung, H.; Oh, Y.
\newblock Towards better explanations of class activation mapping.
\newblock In Proceedings of the Proceedings of the IEEE/CVF International Conference on Computer Vision,  2021, pp. 1336--1344.

\bibitem[Fel \em{et~al.}(2021)Fel, Cad{\`e}ne, Chalvidal, Cord, Vigouroux, and Serre]{fel2021look}
Fel, T.; Cad{\`e}ne, R.; Chalvidal, M.; Cord, M.; Vigouroux, D.; Serre, T.
\newblock Look at the variance! efficient black-box explanations with sobol-based sensitivity analysis.
\newblock {\em Advances in Neural Information Processing Systems} {\bf 2021}, {\em 34},~26005--26014.

\bibitem[Koker \em{et~al.}(2021)Koker, Mireshghallah, Titcombe, and Kaissis]{koker2021u}
Koker, T.; Mireshghallah, F.; Titcombe, T.; Kaissis, G.
\newblock U-noise: Learnable noise masks for interpretable image segmentation.
\newblock In Proceedings of the 2021 IEEE International Conference on Image Processing (ICIP). IEEE,  2021, pp. 394--398.

\bibitem[Nielsen \em{et~al.}(2022)Nielsen, Dera, Rasool, Ramachandran, and Bouaynaya]{nielsen2022robust}
Nielsen, I.E.; Dera, D.; Rasool, G.; Ramachandran, R.P.; Bouaynaya, N.C.
\newblock Robust explainability: A tutorial on gradient-based attribution methods for deep neural networks.
\newblock {\em IEEE Signal Processing Magazine} {\bf 2022}, {\em 39},~73--84.

\bibitem[Vinogradova \em{et~al.}(2020)Vinogradova, Dibrov, and Myers]{vinogradova2020towards}
Vinogradova, K.; Dibrov, A.; Myers, G.
\newblock Towards interpretable semantic segmentation via gradient-weighted class activation mapping (student abstract).
\newblock In Proceedings of the Proceedings of the AAAI conference on artificial intelligence,  2020, Vol.~34, pp. 13943--13944.

\bibitem[Cordts \em{et~al.}(2016)Cordts, Omran, Ramos, Rehfeld, Enzweiler, Benenson, Franke, Roth, and Schiele]{Cordts2016Cityscapes}
Cordts, M.; Omran, M.; Ramos, S.; Rehfeld, T.; Enzweiler, M.; Benenson, R.; Franke, U.; Roth, S.; Schiele, B.
\newblock The Cityscapes Dataset for Semantic Urban Scene Understanding.
\newblock In Proceedings of the Proc. of the IEEE Conference on Computer Vision and Pattern Recognition (CVPR),  2016.

\bibitem[Nasrallah \em{et~al.}(2022)Nasrallah, Samhat, Shi, Zhu, Faour, and Ghandour]{nasrallah2022lebanon}
Nasrallah, H.; Samhat, A.E.; Shi, Y.; Zhu, X.X.; Faour, G.; Ghandour, A.J.
\newblock Lebanon Solar Rooftop Potential Assessment Using Buildings Segmentation From Aerial Images.
\newblock {\em IEEE Journal of Selected Topics in Applied Earth Observations and Remote Sensing} {\bf 2022}, {\em 15},~4909--4918.

\bibitem[Chattopadhay \em{et~al.}(2018)Chattopadhay, Sarkar, Howlader, and Balasubramanian]{gradcampp}
Chattopadhay, A.; Sarkar, A.; Howlader, P.; Balasubramanian, V.N.
\newblock Grad-CAM++: Generalized Gradient-Based Visual Explanations for Deep Convolutional Networks.
\newblock In Proceedings of the 2018 IEEE Winter Conference on Applications of Computer Vision (WACV),  2018, pp. 839--847.
\newblock {\url{https://doi.org/10.1109/WACV.2018.00097}}.

\end{thebibliography}

\PublishersNote{}
\end{adjustwidth}
\end{document}